\documentclass[11pt]{article}
\pdfoutput=1

\usepackage[review]{ACL2023}

\usepackage{times}
\usepackage{latexsym}

\usepackage[T1]{fontenc}

\usepackage[utf8]{inputenc}

\usepackage{microtype}
\usepackage{graphicx}
\usepackage{tabularx}
\usepackage{multirow}
\usepackage{amsmath}

\usepackage{inconsolata}
\usepackage{caption}
\title{Dial-insight: Fine-tuning Large Language Models with High-Quality Domain-Specific Data Preventing Capability Collapse}

\author{ Jianwei Sun, Chaoyang Mei, Linlin Wei, Kaiyu Zheng, Na Liu, Ming Cui, Tianyi Li\\
         Beike Inc., Beijing, China \\ \{sunjianwei006, cuiming001\}@ke.com}

\begin{document}
\maketitle
\begin{abstract}

The efficacy of large language models (LLMs) is heavily dependent on the quality of the underlying data, particularly within specialized domains. A common challenge when fine-tuning LLMs for domain-specific applications is the potential degradation of the model's generalization capabilities. To address these issues, we propose a two-stage approach for the construction of production prompts designed to yield high-quality data. This method involves the generation of a diverse array of prompts that encompass a broad spectrum of tasks and exhibit a rich variety of expressions. Furthermore, we introduce a cost-effective, multi-dimensional quality assessment framework to ensure the integrity of the generated labeling data. Utilizing a dataset comprised of service provider and customer interactions from the real estate sector, we demonstrate a positive correlation between data quality and model performance. Notably, our findings indicate that the domain-specific proficiency of general LLMs can be enhanced through fine-tuning with data produced via our proposed method, without compromising their overall generalization abilities, even when exclusively domain-specific data is employed for fine-tuning.

\end{abstract}

\section{Introduction}

As the capabilities of large language models (LLMs) advance, the expectations and requirements for their application in professional domains have escalated\cite{zhao2023survey,achiam2023gpt}. The integration of LLMs into specialized fields to provide expert-level services is an emerging research focus, intending to leverage their potential to generate significant value\cite{Han_Wei_Liu_Qi_2023}. To enhance LLMs' proficiency in specific domains, it is essential to tailor their capabilities to the unique data characteristics and situational demands of these fields. However, specialized domains often encompass a range of complex, open-ended issues that are intricately linked to the general competencies of LLMs, such as service quality assessment, price calculation, and information extraction\cite{Guo_Jiang_Zhang_Li_Wang_Wang_2023}. If an LLM's general capabilities deteriorate during domain-specific enhancement, each novel problem within the domain necessitates further model refinement, which is an impractical and costly approach. Consequently, domain-specific LLMs must also retain robust general capabilities to address open-ended domain challenges effectively\cite{Jalil_2023,zhang2024scaling}. 
High-quality data is a critical factor in addressing these challenges.\cite{qiu2024wanjuan}

In the quest for high-quality data, scholars have explored various methodologies spanning data source collection, task instruction formulation, and data quality assessment. Chathome\cite{wen2023chathome} gathered a home decoration dataset from diverse sources including books, websites, and repositories to enhance general-purpose LLMs' domain-specific capabilities. However, the data's open-source provenance may limit domain expertise compared to commercial sources. Moreover, without multi-task pre-training (MIP) \cite{Zeng_GLM-130B,T5}, fine-tuning the language model with this data significantly diminishes the base model's general capabilities while enhancing domain-specific skills. CSDS\cite{lin2021csds} developed a Chinese customer service dialogue summary dataset and investigated algorithmic approaches using conventional text summarization techniques. Nevertheless, this dataset is limited to summarizing Q\&A pairs and lacks evaluation of the model's multi-tasking proficiency. In task instruction construction, methods integrating instructions or prompts to refine language model outputs are well-established. \cite{Self-Instruct} introduced SELF-INSTRUCT while WizardLM\cite{WizardLM} proposed an instruction-evolution algorithm, both enhancing instruction construction efficiency. Regarding data quality evaluation, Llama2 et al.\cite{touvron2023llama,bai2023qwen} repeatedly emphasized data quality's impact on LLMs' capabilities. To address building high-quality data through data quality evaluation, \cite{li2023quantity} proposed Instruction Following Difficulty (IFD), enabling models to achieve fine-tuning effects with only 5\%-10\% of original data through micro-tuning. However, this quality index is complex and costly to compute. Accurately assessing data quality at low cost for entire training sets remains challenging.

\begin{figure*}[h]
  \centering
  \includegraphics[width=\linewidth]{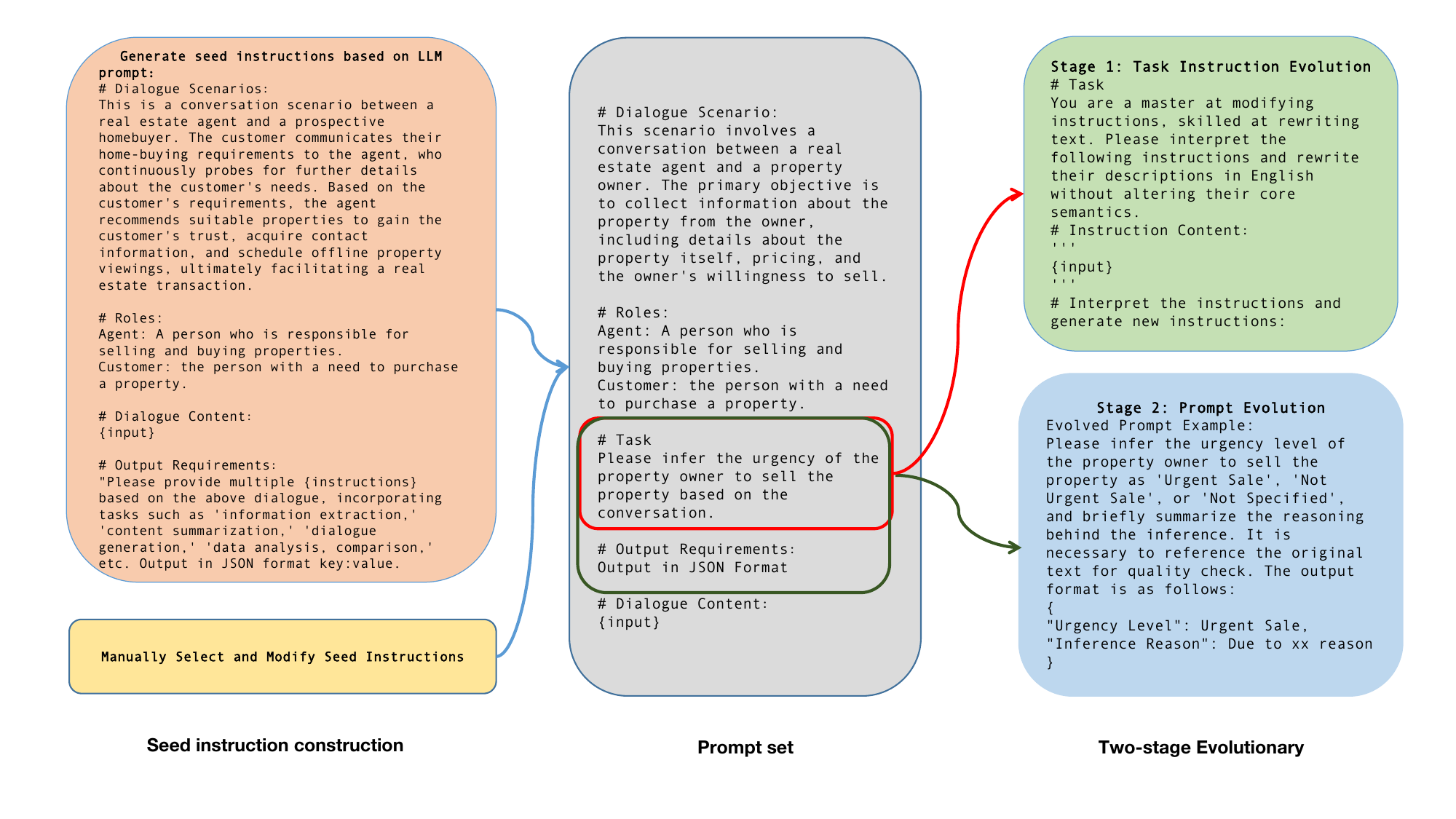}
  \caption{The two-stage Prompt evolution method.}
  \label{Fig:two-satge}
\end{figure*}

This paper introduces a method for producing high-quality data in specialized domains, drawing on an analysis of dialogue data between real estate service providers and customers. We also propose a set of cost-effective, multi-dimensional data quality evaluation metrics. Our experiments demonstrate that the high-quality data generated using this method not only bolsters the professional capabilities of general LLMs but also ensures that the general abilities of domain models do not deteriorate. Compared to existing research, our approach to fine-tuning LLMs with high-quality domain-specific data exhibits three distinct features:

\textbf{1) More intricate instruction construction.} We propose a two-stage evolution method for constructing prompts that integrate domain-specific information, character details, and task directives. These prompts, crafted using our method, exhibit greater complexity and richness than those produced by general instruction evolution algorithms.

\textbf{2) A cost-effective, comprehensive data quality evaluation system.} Diverging from single data quality assessments, we have devised a multi-dimensional, cost-effective comprehensive quality evaluation system to objectively gauge data quality. This system evaluates the entire training dataset's quality based on prompts, inputs, and targets, considering four dimensions: richness, complexity, redundancy, and label quality. Through extensive comparative experiments, we have corroborated that the data quality evaluation results obtained with this system positively correlate with the model training outcomes of the corresponding data, providing interpretability for the model's optimization direction.

\textbf{3) Preservation of general abilities during domain-specific fine-tuning.} Prior research has acknowledged that enhancing domain capabilities often precipitates a decline in the model's general abilities, necessitating the inclusion of general data to maintain overall proficiency. In contrast, our high-quality domain data does not lead to a degradation in the model's general abilities, even when fine-tuning is conducted exclusively with domain data. Our experiments on domain-specific conversation analysis with varying data qualities reveal that as data richness and quality improve, the domain-tuned model's performance on domain-specific tasks significantly enhances without the addition of general multi-task data, and the model's inherent general abilities, such as linguistic proficiency in English and Chinese and general multi-tasking capabilities, remain intact.

\section{Two-stage Evolutionary Prompt Data Production Method}

In the domain of command construction and usage, significant disparities exist between general commands and those tailored to specific vertical domains. The primary distinctions are as follows:

\begin{itemize}
\item \textbf{Instruction Complexity:} Commands within a vertical domain necessitate a multifaceted construction process that considers various elements such as business scenarios, role-specific tasks, and stringent task requirements. Consequently, the complexity of these instructions surpasses that of general command inputs.
\item \textbf{Domain-Specific Task Instructions:} Unlike general commands, instructions in specialized fields are articulated with greater precision and often encompass numerous detailed sub-items. These instructions are characterized by their professional tone and the imperative of high reliability.
\end{itemize}

Given these differences, the evolution of instruction algorithms for general datasets encounters several limitations:

\begin{itemize}
\item \textbf{Uncontrollable Evolution:} General-purpose instruction algorithms evolve based on the instructions themselves, without incorporating the full context of extended text dialogues. This can result in uncontrolled algorithmic evolution when the model's iterative prompts are combined with the complete text input.
\item \textbf{Constrained Instruction Complexity:} Evolutionary algorithms that focus solely on the instructions, without integrating business requirements and the full context of the instructions within the business scenario, yield a limited complexity. The resulting instructions often diverge significantly from those employed in actual business contexts.
\item \textbf{Specialization of Prompt Construction:} The automated evolution of prompts by large-scale models, based on seed tasks, frequently produces prompts that differ markedly in style and complexity from those crafted by human experts in real-world business settings. This discrepancy can diminish the practical utility of the generated prompts.
\end{itemize}

To address these issues, we propose a two-stage Prompt Evolution Method predicated on instruction evolution. The methodological framework is depicted in Fig.\ref{Fig:two-satge}.

\subsection{Stage 1. Task Instruction Evolution}

\textbf{Step 1. Seed Instruction Generation: }
In the initial stage, we draw upon the self-instructive paradigm to create a diverse set of seed instructions for a given task. These instructions are crafted by incorporating multifaceted information from the task's scene, characters, and dialogue, thereby ensuring that the generated instructions are well-aligned with the task's context. This approach marks a departure from traditional methods that focus solely on the task, as it introduces a higher degree of domain specificity into the instruction generation process.

Upon generating a substantial corpus of seed instructions, we recognize the potential for issues such as low validity and semantic redundancy. To mitigate these concerns, we employ a manual data curation process. This involves a thorough review of the seed instructions, during which high-quality instructions are retained and those of lower quality are discarded.

\textbf{Step 2. Task Instruction Evolution: }
Following the manual screening, we are left with a collection of seed instructions that, while preliminary, may lack the desired level of complexity. To address this, we introduce a semi-automated instruction evolution method that combines the capabilities of a large language model with an evolutionary algorithm. The process unfolds as follows:

\begin{itemize}
\item[]
\textbf{Step 2.1. Automated Instruction Enhancement: }
Utilizing the large language model in conjunction with an evolutionary algorithm, we automatically rewrite the seed instructions to augment their complexity and richness. This step aims to evolve the instructions beyond their initial form, introducing variations that could potentially improve their effectiveness.

\textbf{Step 2.2. Manual Screening of Evolved Instructions: }
The evolved instructions are then subjected to a manual screening process. High-quality instructions are identified based on the following criteria: they must be easily understandable and unambiguous, requiring no manual modifications; and they should offer a more detailed and improved task description compared to the original seed instruction.

\textbf{Step 2.3. Human-Guided Instruction Refinement: }
Finally, we engage in a human-guided refinement process, where the selected high-quality instructions are expanded or rewritten. The objective of this step is to correct any semantic distortions or expression issues introduced during the automated evolution phase, thereby optimizing the instructions for normative clarity and coherence.
\end{itemize}
Through the implementation of this two-step methodology, we have successfully generated a set of dialogue analysis instructions characterized by a notable degree of diversity and complexity. This refined set of instructions is poised to enhance the performance of large language models in executing domain-specific tasks.

\subsection{Stage 2. Prompt Evolution}

\begin{figure*}[h]
  \centering
  \includegraphics[width=\linewidth]{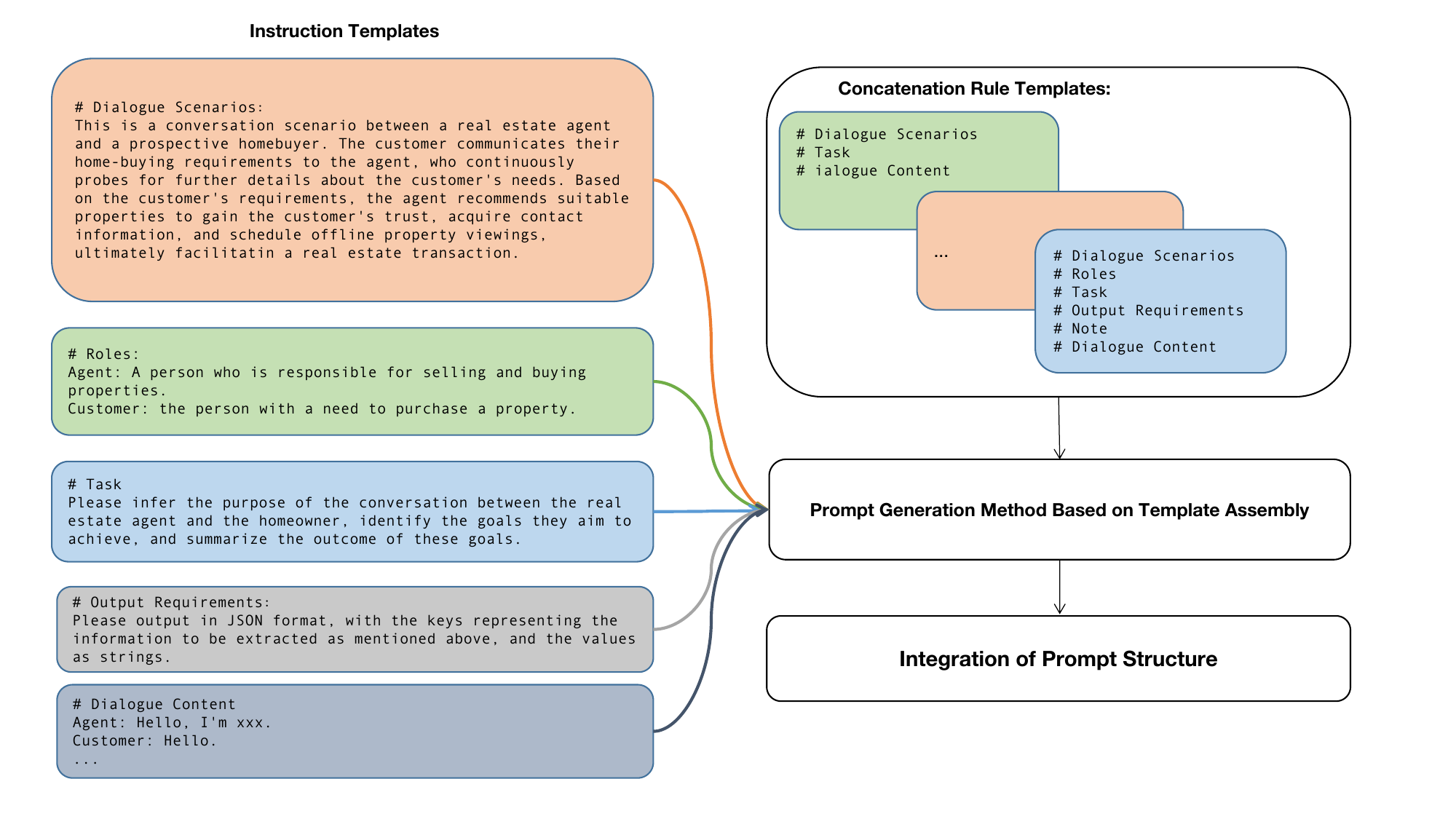}
  \caption{Stage 2: Prompt Evolution method}
  \label{Fig:stage2}
\end{figure*}

The proposed methodology is designed to emulate the intricacies of professional task scenarios, where prompt descriptions are often an intricate fusion of format stipulations and task directives. This is reflective of the expectations and practices of users engaging with sophisticated models, who are accustomed to prompts that harmoniously integrate format guidelines with illustrative examples. The whole method of constructing Prompts is illustrated in Fig.\ref{Fig:stage2}, and the technical process is briefly described as follows:

\textbf{Step 1. Design of Multiple Templates.} We initiate by crafting a diverse array of templates, encompassing various scene descriptors and task instruction narratives. This foundational step ensures that the prompts are grounded in a rich context, catering to a wide range of potential scenarios.

\textbf{Step 2. Assembly-Based Prompt Generation.} Subsequently, we employ a content-filling strategy to randomly amalgamate the sub-components derived from the initial templates. This assembly results in a comprehensive prompt, as depicted in Fig.\ref{Fig:stage2}. The primary merit of this technique lies in its straightforwardness and the ease with which it can be scaled. Nonetheless, it is not without its limitations. The resultant templates, while functionally complete, exhibit a degree of rigidity in the coherence of their constituent parts.

\textbf{Step 3. Enhancement of Prompt Structure.} To surmount the aforementioned limitation, we introduce an additional step that refines the prompt structure. Drawing inspiration from human-written examples, we first integrate the output format, overarching guidelines, and task directives to reformulate the instruction narrative. Thereafter, we apply the previously established template splicing method to seamlessly blend background context, character exposition, dialogue elements, and the revised instruction narrative into a cohesive Prompt.

\subsection{Data Label Generation}

To mitigate the challenge of data distribution imbalance and to prioritize the development of specific model capabilities during training, we adopt a weighted random sampling strategy for the selection of task instructions in the data label generation process. This method ensures a balanced distribution of training samples and fosters a diverse range of task instructions and expressions. Additionally, it allows for targeted enhancement of scene-specific capabilities. This nuanced approach to data curation is crucial for the advancement of robust and versatile models in natural language processing tasks. The method is outlined as follows:

\begin{itemize}
\item[]
\textbf{Step 1.} Scene-based conversation data weighting-selecting conversation data randomly based on the importance of the scene and task.

\textbf{Step 2.} Generate the full task description Prompt based on the two-stage Prompt evolution method.

\textbf{Step 3.} Based on the general model, automatically produce corresponding data labels.
\end{itemize}

\section{Experiment \& Analysis}
\subsection{Data Set Construction}

In the real estate domain, there are numerous scenarios for dialogue, with a vast number of dialogues generated from various sources and scenarios. We conducted data sorting based on the nature of the business and role behavior to organize the raw data.

Based on the nature of the actual transaction business, the dialogue scene can be roughly divided into four scenarios: "buying the house", "selling the house", "renting the house", and "Home improvement".

According to the source of data, the data can be divided into 10 different types, including "call recordings", "large screen recordings", "smart badge recordings", "VR recordings", "PAD recordings", "mobile phone recordings", "single text records", "IM conversation messages", "corporate WeChat group messages", and "personal corporate WeChat messages".

\textbf{Task Type Classification:} In the scene, service providers include various roles such as "real estate agents", "home designers", "house rental cloud managers", "lease asset managers", "customer managers", and "home advisors" who perform daily tasks that can generate various tasks, such as "communication between agents and owners", "new customer sources", etc. Based on the actual business content of the service providers, we have collected and summarized the main task types as shown in Table \ref{table: task type}

During the experimental process, we continuously improved and perfected the data production method, and constructed and perfected multiple versions of the real estate dialogue insight dataset, including "Dial-insight-train-v0.5", "Dial-insight-train-v1.0" for training, and the "Dial-insight-test" multi-task testing set. The construction methods of these datasets are shown in Table \ref{table:Generation Methods}.

\begin{table*}[htb]
\caption{ Data Generation Methods}
\label{table:Generation Methods}
\resizebox{\textwidth}{!}{
\begin{tabular}{lllll}
\hline
Data Version &
  Data Type &
  \begin{tabular}[c]{@{}l@{}}Method for \\ Prompt Generation\end{tabular} &
  \begin{tabular}[c]{@{}l@{}}Method for \\ Data Label Generation\end{tabular} &
  Data Security \\ \hline
\begin{tabular}[c]{@{}l@{}}Dial-insight\\ -train-v0.5\end{tabular} &
  \begin{tabular}[c]{@{}l@{}}training \\ set\end{tabular} &
  \begin{tabular}[c]{@{}l@{}}Manually craft a fixed \\ number of prompts.\end{tabular} &
  \multirow{3}{*}{\begin{tabular}[c]{@{}l@{}}Leveraging a generic model \\ for the pre-annotation of data, \\ subsequent refinements are \\ made through manual methods \\ to enhance the outcomes of the \\ model.\end{tabular}} &
  \multirow{3}{*}{\begin{tabular}[c]{@{}l@{}}Removing sensitive \\ user information \\ such as names, phone \\ numbers, account \\ numbers, passwords\\ , etc.\end{tabular}} \\ \cline{1-3}
\begin{tabular}[c]{@{}l@{}}Dial-insight\\ -train-v1.0\end{tabular} &
  \begin{tabular}[c]{@{}l@{}}training \\ set\end{tabular} &
  \begin{tabular}[c]{@{}l@{}}Manually revise the \\ prompt based on template \\ concatenation.\end{tabular} &
   &
   \\ \cline{1-3}
\begin{tabular}[c]{@{}l@{}}Dial-insight\\ -test\end{tabular} &
  \begin{tabular}[c]{@{}l@{}}multi-task \\ test set\end{tabular} &
  \begin{tabular}[c]{@{}l@{}}Randomly generating part of\\ the sample by the 2-stage\\ prompts production method.\\ based on multi-task type.\end{tabular} &
   &
   \\ \hline
\end{tabular}
}
\end{table*}

\begin{figure*}
    \centering
    \includegraphics[width=\linewidth]{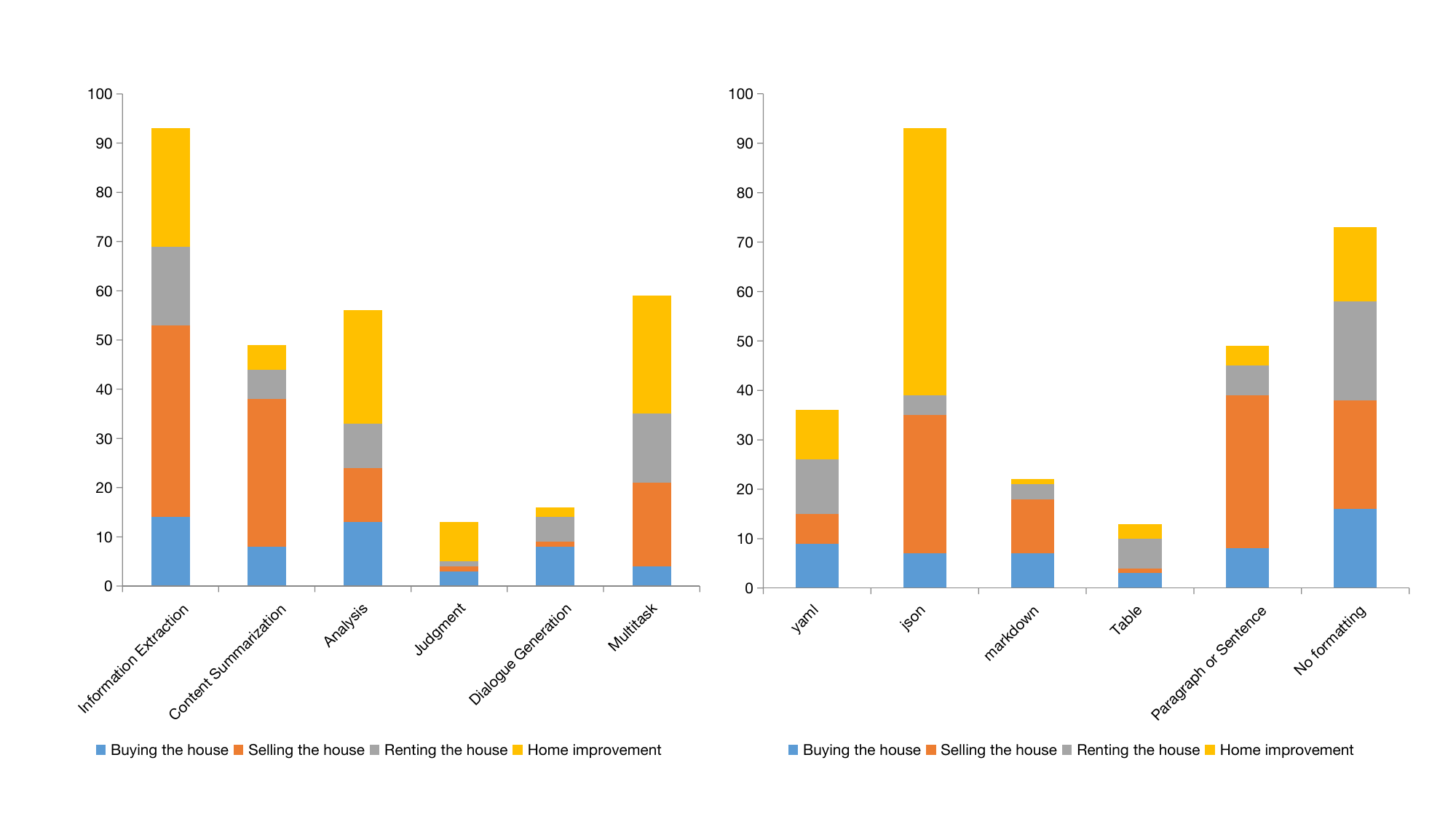}
    \caption{The distribution statistics of task types (a) and the distribution statistics of task output formats (b) in the test dataset. Please note that the units on the vertical axis represent the number of test data samples.}
    \label{Fig:testdata}
\end{figure*}

\begin{table*}[htb]
\caption{Comparison of the impact of training data size on training effectiveness}
\label{table:Comparision train size}
\centering
\begin{tabular}{cccc}
\hline
{\color[HTML]{333333} Model-name} &
  Base model &
  {\color[HTML]{333333} \begin{tabular}[c]{@{}c@{}}Dial-insight-train-v0.5 \\ Training Sample Count\end{tabular}} &
  {\color[HTML]{333333} \begin{tabular}[c]{@{}c@{}}Dial-insight-test\\ 10 score\end{tabular}} \\ \hline
{\color[HTML]{333333} GPT3.5}           & -                                    & {\color[HTML]{333333} -}     & {\color[HTML]{333333} 7.24} \\
{\color[HTML]{333333} Qwen-14B-Chat}    & -                                    & {\color[HTML]{333333} -}     & 6.13                        \\
{\color[HTML]{333333} Dial-insight-14B-Chat-only-Domain} & {\color[HTML]{333333} Qwen-14B-Chat} & {\color[HTML]{333333} 7000}  & {\color[HTML]{333333} 6.25} \\
{\color[HTML]{333333} Dial-insight-14B-Chat-only-Domain} & {\color[HTML]{333333} Qwen-14B-Chat} & {\color[HTML]{333333} 40000} & {\color[HTML]{333333} 6.77} \\ \hline
\end{tabular}
\end{table*}

\subsection{Model training}

\textbf{Baseline:} We fine-tune the large language model based on Qwen-14B-chat\cite{bai2023qwen} by incorporating the dial-insight training set for conversation insight. Then, we compare the performance of the conversation insight model with GPT-4\cite{achiam2023gpt}, GPT-3.5\cite{gpt3.5}, Mixtral-8x7B-Instruct-v0.1(MoE)\cite{jiang2024mixtral}, Qwen-72B-Chat\cite{bai2023qwen}, Yi-34B-Chat\cite{ai2024yi}, and Qwen-14B-Chat on both vertical task datasets and open-source general task benchmarks.

\textbf{Training parameters:} epoch 3, lr: 2e-5, GPUs: 8, batch size 128, full-parameter fine-tuning

\subsection{Data Comprehensive Quality Evaluation}

We have designed a comprehensive quality evaluation system that assesses data quality and selects high-quality data for model domain tuning. This system consists of six indicators: prompt complexity, input complexity, data richness, data redundancy, and data label quality.

The Prompt Complexity method is as (1):

\begin{equation}
C_{Prompt}= \sum\limits_0^n (L_{
P})/n
\end{equation}
Where the \(L_{P}\) represents the token length per prompt, the \(n\) represents the total number of the dataset.

The Input Complexity method is as (2):

\begin{equation}
C_{input}= \sum\limits_1^n (\frac{1}{3} L_{
C}+ \frac{2}{3} L_{
P})/n
\end{equation}
Where the \( L_{C}\) represents the token length per dialogue.

A brief description of the method for calculating the richness metric is as follows. Initially, data undergo text vector representation using the Qwen-14B-Chat model tokenizer. Clustering is then performed utilizing a community detection algorithm. The core concept here involves determining the cosine similarity between two samples to assess their likeness; when this similarity exceeds a certain threshold (set at 0.82 for this study, following experimental validation), samples are considered similar and thus eligible for inclusion in the selection pool; otherwise, they are excluded from consideration. Once the number of samples in the selection pool reaches a predetermined count, diversity filtering is deemed complete. The Richness method is as (3):

\begin{equation}
    Ri =  \sum\limits_1^N(C_{i} \cdot N_{Top_{n }} )/n,(n>=1,N>=1)
\end{equation}
Where the \(N\) represents the number of centroids after clustering, the \(C_{i}\) is the \(i\)-th cluster center of the dataset, and the \(N_{Top_{n }} \) is the top n quantity for each category.

The Data Redundancy is as (4):

\begin{equation}
    Re =  1-Ri
\end{equation}

The Data Label Quality method is as (5):

\begin{equation}
    Q = \frac{N_{x}}{N_{total}} \text{, (}x=high,medium, low\text{)}
\end{equation}
Where the \(N_{x}\) represents the total number of samples belonging to a certain quality category.

Comparing the two data versions in Table \ref{table:data quality}, it is evident that there are significant differences in data quality between v0.5, and v1. Through the two-stage prompt Evolution iterations, the richness of the training dataset has greatly improved, and the data redundancy indicator has noticeably decreased. Overall, the quality of the data has seen significant improvement.

\subsection{Model performance evaluation}

\begin{table*}[htb]
\caption{The Impact of Data Quality on Model Performance}
\label{table:data quality}
\resizebox{\textwidth}{!}{
\begin{tabular}{cccccccccccc}
\cline{1-12}
  {\color[HTML]{333333} } &
  \multicolumn{7}{c}{Data Quality Metrics} &  
  \multicolumn{3}{c}{{\color[HTML]{333333} Dial-insight-14b-only-Domain}} \\ \cline{3-12}
 &
  {\color[HTML]{333333} } &
   &
   &
   &
   &
  \multicolumn{3}{c}{Label Quality} &
  {\color[HTML]{333333} } &
  {\color[HTML]{333333} } &
  {\color[HTML]{333333} } \\
\multirow{-3}{*}{Data Name} &
  \multirow{-3}{*}{{\color[HTML]{333333} \begin{tabular}[c]{@{}c@{}}Full session\\ Data Volume\end{tabular}}} &
  \multirow{-2}{*}{Richness} &
  \multirow{-2}{*}{Input Complexity} &
  \multirow{-2}{*}{Prompt Complexity} &
  \multirow{-2}{*}{Repetition Rate} &
  High &
  Medium &
  Low &
  \multirow{-2}{*}{{\color[HTML]{333333} dial-insight-test}} &
  \multirow{-2}{*}{{\color[HTML]{333333} \begin{tabular}[c]{@{}c@{}}c-veval\\ 0-shot\end{tabular}}} &
  \multirow{-2}{*}{{\color[HTML]{333333} \begin{tabular}[c]{@{}c@{}}c-veval\\ 5-shot\end{tabular}}} \\ \hline
v0.5 &
  40000 &
  1.56\% &
  1174.1 &
  1300 &
  98.44\% &
  56.87\% &
  34.20\% &
  8.93\% &
  {\color[HTML]{333333} 6.77} &
  {\color[HTML]{333333} 65.3} &
  {\color[HTML]{333333} 67.9} \\
v1.0 &
  40000 &
  {\color[HTML]{333333} 69.74\%} &
  {\color[HTML]{333333} 1078.2} &
  {\color[HTML]{333333} 273} &
  {\color[HTML]{333333} 30.26\%} &
  97.00\% &
  3.00\% &
  0.00\% &
  {\color[HTML]{333333} 8.24} &
  {\color[HTML]{333333} 66.57} &
  {\color[HTML]{333333} 67.9} \\
{\color[HTML]{333333} v1.0} &
  {\color[HTML]{333333} 80000} &
  {\color[HTML]{333333} 73.72\%} &
  {\color[HTML]{333333} 1401} &
  {\color[HTML]{333333} 281} &
  {\color[HTML]{333333} 26.74\%} &
  95.67\% &
  2.67\% &
  1.66\% &
  {\color[HTML]{333333} 8.12} &
  {\color[HTML]{333333} 67.86} &
  {\color[HTML]{333333} 68.8} \\ \hline
  
\end{tabular}
}
\end{table*}

\begin{table*}[htb]
\caption{Multi-Model Performance Comparison}
\label{table:multi-model}
\resizebox{\textwidth}{!}{
\begin{tabular}{ccccccccccc}
\cline{1-11}
\multirow{2}{*}{model name} &
  \multicolumn{3}{c}{Training Sample Count} & 
  \multirow{2}{*}{dial-insight-test} &
  \multirow{2}{*}{\begin{tabular}[c]{@{}c@{}}c-eval \\ (0-shot)\end{tabular}} &
  \multirow{2}{*}{\begin{tabular}[c]{@{}c@{}}c-eval\\ (5-shot)\end{tabular}} &
  \multirow{2}{*}{\begin{tabular}[c]{@{}c@{}}mmlu\\ (0-shot)\end{tabular}} &
  \multirow{2}{*}{\begin{tabular}[c]{@{}c@{}}mmlu\\ (5-shot)\end{tabular}} &
  \multirow{2}{*}{\begin{tabular}[c]{@{}c@{}}cmmlu\\ (0-shot)\end{tabular}} &
  \multirow{2}{*}{\begin{tabular}[c]{@{}c@{}}cmmlu\\ (5-shot)\end{tabular}} \\
 &
  Domain &
  \begin{tabular}[c]{@{}c@{}}Chinese \\ General\end{tabular} &
  \begin{tabular}[c]{@{}c@{}}English \\ General\end{tabular} &
   &
   &
   &
   &
   &
   &
   \\
\cline{1-11}
GPT4-turbo &
  - &
  - &
  - &
  8.64 &
  - &
  - &
  - &
  - &
  - &
  - \\
GPT3.5 &
  - &
  - &
  - &
  6.17 &
  - &
  - &
  - &
  - &
  - &
  - \\
Mixtral-8x7B-Instruct-v0.1 &
  - &
  - &
  - &
  6.18 &
  49.03 &
  53.05 &
  49.33 &
  51.95 &
  66.56 &
  69.29 \\
Yi-34B-Chat &
  - &
  - &
  - &
  6.19 &
  74.74 &
  79.12 &
  75.88 &
  81.19 &
  67.54 &
  73.7 \\
Qwen-72B-Chat &
  - &
  - &
  - &
  7.18 &
  54.31 &
  83.28 &
  62.22 &
  83.03 &
  58.76 &
  76.25 \\
Qwen-14B-Chat &
  - &
  - &
  - &
  6.13 &
  62.85 &
  69.99 &
  61.82 &
  65.24 &
  67.41 &
  70.56 \\
Dial-insight-14b-only-Domain &
  70000 &
  0 &
  0 &
  8.32 &
  65.82 &
  68.5 &
  61.38 &
  62.9 &
  66.63 &
  68.94 \\
Dial-insight-14b-only-Domain &
  80000 &
  0 &
  0 &
  8.12 &
  67.86 &
  68.8 &
  61.48 &
  63.36 &
  67.66 &
  69.16 \\
Dial-insight-14b-Chat &
  70000 &
  50000 &
  10000 &
  \textbf{8.38} &
  66.05 &
  67.09 &
  60.03 &
  61.62 &
  64.7 &
  66.44 \\ \hline

\end{tabular}
}
\end{table*}

Test dataset: we use two Chinese open-source general task evaluation sets, one English open-source general task evaluation set, and one Chinese real estate dialogue analysis field task evaluation set to evaluate the model's ability for multiple large models.

\begin{itemize}
\item[]
\textbf{C-Eval} \cite{huang2024ceval} C-Eval is a Chinese evaluation dataset spanning 52 diverse disciplines. We report 0-shot and 5-shot results.

\textbf{CMMLU} \cite{li2023cmmlu} CMMLU is designed for assessing language understanding capabilities in Chinese. We report 0-shot and 5-shot results.

\textbf{MMLU} \cite{mmlu} Massive Multi-task Language Understanding is designed for measuring language understanding capabilities. We report 0-shot and 5-shot results. 

\textbf{Dial-insight-test} dataset is constructed based on 286 conversations between real estate service providers and clients, covering multiple tasks such as information extraction, content summarization, reasoning ability, speech generation, data analysis, and comparison. The distribution statistics of task types in the test set and the distribution statistics of task output formats are shown in Fig \ref{Fig:testdata}.
\end{itemize}

Evaluation methods: It is important to note that we are asking the GPT-4 model to reference the labeled answers in the test set and evaluate the output results for the model in a specific domain. The prompt for scoring is as follows:

\subsection{Experimental results}

We fine-tuned the Qwen-14B-chat model with v0.5 data of different scales, and the results are shown in Table \ref{table:Comparision train size}. By comparing, we can see that increasing the data scale can improve the model performance. However, the performance improvement of the model using this dataset is limited.

Furthermore, we conducted multiple sets of controlled experiments by enhancing the overall quality of the data and performing comprehensive quality assessments on the training data. As shown in Table \ref{table:data quality}, we found that as the overall quality of the data improved, the effectiveness of the model's vertical domain capabilities significantly strengthened. At the same time, the model's general performance will also improve. It should be noted that when it comes to data scale, bigger is not necessarily better. In practical model training, it is imperative to further investigate the appropriate size of the dataset in order to optimize model performance.

We compared the domain-specific models with multiple open-source general models, and the comparison results are shown in Table \ref{table:multi-model}. Among them, the Dial-insight-14B-Chat model is based on the Qwen-14B-Chat as the base model, trained using Dial-insight training data and some open-source general Chinese and English data. It is a large model for dialogue analysis in the real estate domain obtained after full-parameter fine-tuning. Dial-insight14b-only-Domain model differs from the former in that it only uses dial-insight training data as fine-tuning data during the fine-tuning process and does not use any open-source general data. From Table \ref{table:multi-model}, it can be seen that after strengthening with dial-insight training data, compared to the base model, there is a significant improvement in domain-specific scenarios. At the same time, after the model is strengthened using only Dial-insight data for general models, including language abilities, there is no significant collapse in the original general capabilities.

Through the above experiment, we observed that, given the significant improvement in data quality indicators such as the richness of vertical domain data and label quality, the model, after being fine-tuned with vertical domain data only, demonstrates a noticeable enhancement in its vertical domain capabilities. Importantly, this enhancement does not result in a significant degradation of the model's original general capabilities.

\section{Conclusion}

This paper proposes a high-quality domain data construction production method, which mainly includes a two-stage prompt evolution method and a data labeling method based on large models. Then, based on a comprehensive evaluation system for data quality, the quality of the produced data is monitored. Finally, based on the production data, large models are trained with full parameters in the vertical domain, and the effectiveness of the high-quality domain data in improving the performance of large models in the vertical domain is verified through multiple model comparison experiments, while not causing the collapse of the original general ability of the large models.

\bibliography{main}
\bibliographystyle{acl_natbib}

\appendix

\begin{table*}[hbt]
\caption{Task Type Statistics}
\label{table: task type}

\resizebox{\textwidth}{!}{
\begin{tabular}{|l|l|}

\hline
Task type                    & Task Requirements           \\ \hline
Information Extraction       & \begin{tabular}[c]{@{}l@{}}1. Identify the issues that the service provider needs to pay attention to \\ based on the dialogue content.\\ 2. Extract the needs, basic information, attitudes, emotions, intentions, \\ profanity, and language of the served party.\\ ...\end{tabular}\\ \hline
Content Summarization        & \begin{tabular}[c]{@{}l@{}}1. Summarize and analyze the content of the dialogue.\\ 2. Summarize the service provider's service process.\\ 3. Summarize the customer's communication intention transformation \\ process.\\ ...\end{tabular}                  \\ \hline
Reasoning Ability            & \begin{tabular}[c]{@{}l@{}}1. Predict potential customer complaints, churn risk, and other related \\ information based on existing conversations, and issue warnings \\ accordingly.\\ 2. Provide communication suggestions to the service provider based on\\  communication trends to achieve goals.\\ 3. Outline the reasoning process and willingness for each task.\\ 4. Summarize the inference with prior knowledge.\\ ...\end{tabular}                      \\ \hline
Dialogue Generation          & \begin{tabular}[c]{@{}l@{}}1. Generate corresponding solution proposals for the issues mentioned\\  in the conversation.\\ 2. Assist the service provider in enriching their language based on \\ their needs.\\ 3. Service Provider Assistant - Knowledge Integration.\\ 4. Generate copywriting for property information.\\ 5. Solve customer problems or assist service providers in solving \\ customer problems based on known information such as knowledge \\ base and property information.\\ ...\end{tabular} \\ \hline
Data Analysis and Comparison & \begin{tabular}[c]{@{}l@{}}1. Conduct matching analysis based on property information combined \\ with customer requirements, and complete recommendations.\\ 2. Analysis of property comparisons, neighborhood comparisons, etc.\\ 3. Generate horizontal service reports.\\ 4. Calculate customer retention rate.\\ 5. Evaluate service effectiveness. \\ ...\end{tabular}                     \\ \hline
Multitask Integration        & \begin{tabular}[c]{@{}l@{}}Write a comprehensive instruction that combines various single tasks \\ and allows the model to output multiple results at once.\\ ...\end{tabular}    \\ \hline
\end{tabular}
}
\end{table*}

\end{document}